\documentclass[10pt,twocolumn,letterpaper]{article}

\usepackage{iccv}
\usepackage{times}
\usepackage{graphicx}
\usepackage{amsmath}
\usepackage{authblk}

\usepackage[pagebackref=true,breaklinks=true,letterpaper=true,colorlinks,bookmarks=false]{hyperref}

\iccvfinalcopy 

\def\iccvPaperID{826} 

\ificcvfinal\fi

\graphicspath{{figures/}}
\newcommand{\bftab}{\fontseries{b}\selectfont}

\begin{document}

\title{End-to-end learning potentials for structured attribute prediction}

\author[1]{Kota Yamaguchi}
\author[2,3]{Takayuki Okatani}
\author[4]{Takayuki Umeda}
\author[4]{Kazuhiko Murasaki}
\author[5]{Kyoko Sudo}
\makeatletter
\renewcommand\AB@affilsepx{\hspace{2em}\protect\Affilfont}
\makeatother

\affil[1]{CyberAgent, Inc.}
\affil[2]{Tohoku University}
\affil[3]{RIKEN}
\affil[4]{NTT}
\affil[5]{Toho University}

\maketitle

\begin{abstract}
We present a structured inference approach in deep neural networks for multiple attribute prediction. In attribute prediction, a common approach is to learn independent classifiers on top of a good feature representation. However, such classifiers assume conditional independence on features and do not explicitly consider the dependency between attributes in the inference process. We propose to formulate  attribute prediction in terms of marginal inference in the conditional random field. We model potential functions by deep neural networks and apply the sum-product algorithm to solve for the approximate marginal distribution in feed-forward networks. Our message passing layer implements sparse pairwise potentials by a softplus-linear function that is equivalent to a higher-order classifier, and learns all the model parameters by end-to-end back propagation. The experimental results using SUN attributes and CelebA datasets suggest that the structured inference improves the attribute prediction performance, and possibly uncovers the hidden relationship between attributes.
\end{abstract}

\section{Introduction}

Deep neural networks achieve surprisingly strong performance in various computer vision tasks including detection, segmentation, and attribute recognition. In attribute recognition, typically the goal is to predict multiple semantic labels to the given image, and sometimes this problem formulation is referred automatic image tagging. Attributes can be any visually perceptible concept, such as \emph{outdoor} or \emph{man-made} for a scene image. Many attributes co-exist in a single image, and some are in an exclusive relationship (e.g., \emph{outdoor} vs. \emph{indoor}). In this paper, we consider an approach to explicitly modeling such inter-attribute relationships in the inference process using deep neural networks.

A standard approach to predict multiple labels is to learn independent classifiers at the end of the neural network (e.g., fork architecture~\cite{Zhang2014PANDA:Modeling}). Through end-to-end learning, the trunk of the architecture is expected to learn discriminative internal representation such that the leaves can make good predictions for each concerned attribute assuming attributes are conditionally independent. However, inter-attribute relationship is not explicitly modeled in the feed-forward inference in such fork architecture, and the hope is that the internal representation is discriminative enough to predict all of the attributes. Alternatively, one might be able to learn independent neural networks for each attribute, with a drawback in the scalability for learning multiple huge networks. In this paper, we aim at building a drop-in replacement for conditionally independent classifiers.

\begin{figure*}[t]
  \centering
  \includegraphics[width=\textwidth]{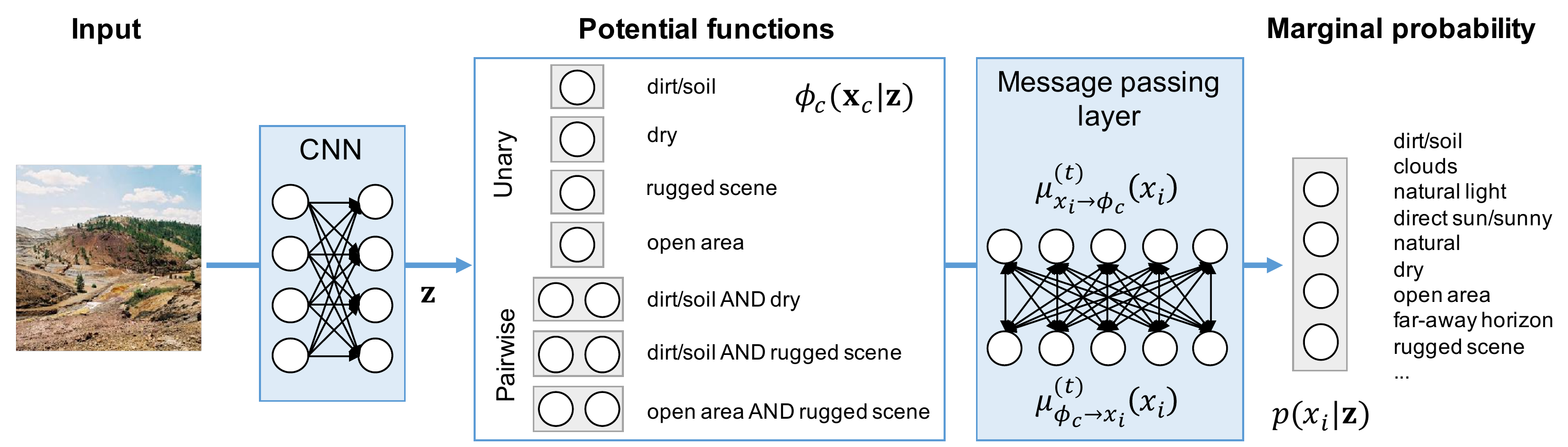}
  \caption{Our structured attribute prediction model. We introduce the sum-product message passing for marginal inference in the end-to-end learnable architecture. The base network predicts potential functions to explicitly model attribute relationships.}
  \label{fig:arch}
\end{figure*}

We attempt to explicitly model the inter-attribute relationship via message passing inference in a feed-forward network, and learn the whole network in an end-to-end manner. Our approach is based on the (approximate) marginal inference by the sum-product algorithm in the conditional random fields (CRF), and we propose to learn sparse potential functions on top of the feature representation produced by deep neural networks. Predicting potential functions by neural networks has an advantage that we can consider arbitrary order relationships in the prediction~\cite{Lin2015DeeplyInference}. For example, in facial attribute recognition, in addition to the probability of \emph{Male} given an image, a potential function can consider \emph{Male and Beard} probability together with the cue from \emph{Male and 5 o'clock shadow} probability to infer the marginal probability of \emph{Male}, thus taking advantage of inter-dependent contexts in the final prediction. Our message passing layer can be implemented as a feed-forward computation in deep networks by unrolled message propagation, and also the whole network can be efficiently learned by stochastic gradient descent (SGD) using standard back propagation. Fig.~\ref{fig:arch} illustrates our model. Given an image, our model predicts potential functions of CRF using convolutional neural networks (CNN), then run joint inference in the message passing layer to solve for the marginal probability. We show in the experiments using two benchmark datasets that our structured inference improves the performance over the independent baseline prediction even if the pairwise potentials are much sparser than a fully-connected dense graph. We also empirically examine properties of the message passing layer in terms of graph structure, propagation steps, and runtime efficiency. Our contributions are summarized in the following:

\begin{itemize}
\item We propose to learn potential functions in the deep neural networks for multiple attribute recognition. We suggest softplus linear function to implement arbitrary order potential function. With potential functions, deep neural networks can explicitly model higher-order relationship between attributes.
\item We propose to use the unrolled sum-product message passing to efficiently run (approximate) marginal inference on a sparse factor graph in a feed-forward network, and also to enable back propagation for end-to-end learning.
\item We empirically show that the structured inference can improve attribute recognition performance with sparse graphical models. We also extensively analyze the effect of graph structure, message propagation steps, and runtime of our message passing layer to derive design principles.
\end{itemize}

\section{Related work} \label{sec:related-work}
There have been several attempts to introduce structured inference in deep neural networks, for example, in semantic segmentation~\cite{Chen2015SemanticCRFs,Ionescu2015MatrixLayers,Lin2015DeeplyInference,Zheng2015ConditionalNetworks,Arnab2016HigherNetworks,Wang2015,Lin2016EfficientSegmentation,Chandra2016FastCRFs,Vemulapalli2016GaussianSegmentation}, pose estimation~\cite{Tompson2014JointEstimation,Wei2016ConvolutionalMachines,Kirillov2016JointOptimization,Chu2016CRF-CNN:Estimation}, stereo~\cite{Knobelreiter2016End-to-EndStereo}, optical flow~\cite{Wang2016ProximalModels}, or action/attribute recognition~\cite{yamaguchi2015mix,Jain2016Structural-RNN:Graphs,Hu2016LearningRelations,Deng2015DeepRecognition,Ramanathan2015LearningImages}. In semantic segmentation, often the mean field approximation is employed~\cite{Chen2015SemanticCRFs,Zheng2015ConditionalNetworks}, where the joint model considers labeling of spatially neighboring pixels. Chu et al.~\cite{Chu2016CRF-CNN:Estimation} use mean field approximation to propagate messages in the feature representation for pose estimation. The mean-field approximation is a popular choice in marginal inference due to computational efficiency~\cite{saito2015transformation,Baque2016PrincipledFields}, though the resulting distribution could be inexact compared to alternative inference methods.

Several inference approaches are reported, for example, by utilizing sampling to learn generic potential functions~\cite{Kirillov2016JointOptimization}, feed-forward message propagation according to label relationships~\cite{Hu2016LearningRelations}, recurrent models to learn the inter-label relationships~\cite{Deng2015DeepRecognition,Deng2015}, or matrix computation in the feed-forward procedure~\cite{Ionescu2015MatrixLayers}. Belanger et al. proposes an energy-based, discriminative approach to learn structure in the deep networks~\cite{Belanger2016StructuredNetworks,Belanger2017End-to-EndNetworks}. In this paper, we try to integrate marginal inference by the sum-product algorithm~\cite{pearl1988probabilistic} in the deep neural networks.

We employ sum-product inference because sparse graphical models are sufficient to improve performance in attribute recognition without incurring prohibitive computational costs. Our model is similar in spirit to Lin et al.~\cite{Lin2015DeeplyInference} where a CNN is utilized to directly predict messages, whereas our model learns potential functions which is somewhat more interpretable. Chen et al. propose an algorithm to integrate both belief propagation and learning in the end-to-end learning~\cite{Chen2015LearningModels}, and we further examine data dependent functions for potentials in a sparse graph. Liu et al.~\cite{Liu2015Multi-ObjectiveLabeling} reports an attempt to learn unary and pairwise potential functions under multi-objective loss, whereas we propose to learn both of the potentials towards the common loss. Attribute recognition has a property that variables are much smaller than other joint problems such as segmentation, but label relations are highly semantic than spatial. We show in the experiment that this problem nature makes sum-product inference in a sparse graphical model suitable for multiple attribute recognition.

Note that, Poon and Domingos propose the sum-product networks (SPN) as alternative deep models to compute marginal probability~\cite{poon2011sum}, and recently an efficient computation of moments on SPN is reported~\cite{Zhao2017EfficientNetworks}. It would be interesting to study in the future how we can take advantage of SPN in standard feed-forward networks.

\section{Structured inference} \label{sec:structured-inference}
To introduce structure in attribute prediction, we consider joint probability over attributes, and view the prediction of each attribute in terms of marginal inference.

\subsection{Conditional random fields} \label{sec:crf}
Let us denote attributes by a vector ${\bf x} = \{ x_0, \cdots, x_N \}$, and a conditional variable by ${\bf z}$ (e.g., feature vector). In CRF, the probability distribution is expressed by the product of potential functions over graph cliques:
\begin{align}
  P({\bf x}|{\bf z}) & = \frac{1}{Z} \prod_c \phi_{c}({\bf x}_c|{\bf z}) \label{eq:crf}
\end{align}
where $Z$ is a normalization constant and $\phi_c$ is the potential function over the clique $c$. We consider computing the conditional marginal distribution:
\begin{align}
p(x_i|\mathbf{z}) = \sum_{\mathbf{x} \backslash x_i} P(\mathbf{x}|\mathbf{z}), \label{eq:marginal_definition}
\end{align}
using the sum-product algorithm. The advantage of using CRF is that we are able to model arbitrary order potential functions $\phi_c({\bf x}_c|{\bf z})$ instead of directly making conditionally independent prediction. Marginal probability naturally incorporates structured relationship between variables in arbitrary order in the probabilistic inference. We can model potential functions in any manner as long as the function returns positive values, for example, using deep neural networks.

\subsection{Sum-product algorithm} \label{sec:sum-product}
The sum-product algorithm computes the marginal probability through message passing on a factor graph. The factor graph is a bipartite graph consisting of factor and variable nodes corresponding to potential functions and variables in Markov Random Fields (MRF). Edges in the factor graph correspond to variable-factor relationships in MRF. The sum-product algorithm computes two types of message propagation in the factor graph:
\begin{align}
  \mu_{\phi_c \rightarrow x}^{(t)}(x) & =
  \sum_{\mathbf{x}_c \backslash x} \phi_c (\mathbf{x}_c|\mathbf{z}) \prod_{m \in \text{n}(\phi_c) \backslash x} \mu^{(t-1)}_{x_m \rightarrow \phi_c}(x_m), \label{eq:message1} \\
  \mu^{(t)}_{x \rightarrow \phi_c}(x) & \propto
  \prod_{l \in \text{n}(x) \backslash \phi_c} \mu^{(t)}_{\phi_l \rightarrow x}(x), \label{eq:message2}
\end{align}
where $t \in \{ 1, \cdots, T \}$ is the iteration step and $\mathrm{n}(\phi_c)$ represents neighbors of $\phi_c$. The message propagation iterates under a pre-determined schedule or until convergence criteria are met.

The final marginal probability is obtained by the product of messages to the variable:
\begin{align}
  p(x_i|\mathbf{z}) & \propto \prod_{c \in \text{n}(x_i)} \mu^{(t)}_{\phi_c \rightarrow x_i}(x_i). \label{eq:marginal}
\end{align}

With an appropriate propagation schedule, the above message passing can compute exact marginal if the graph is a tree. For a general graph with loops, the algorithm can approximately compute the marginal probability under designated message-passing schedule (loopy belief propagation)~\cite{frey1998revolution}. 

\subsection{Message passing layer} \label{sec:message-passing-layer}
\begin{figure}
  \centering
  \includegraphics[width=\columnwidth]{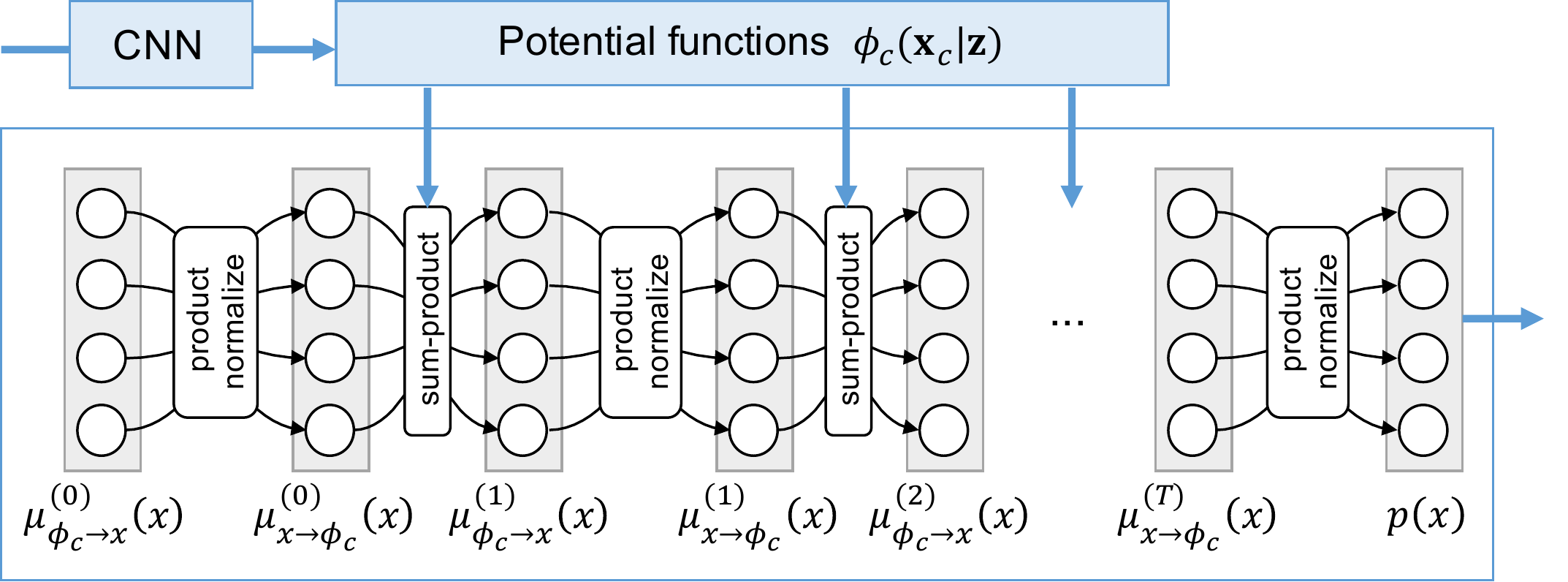}
  \caption{Message passing layer. We implement iterated message propagation by unrolled feed-forward computation.}
  \label{fig:message-passing-layer}
\end{figure}

In this paper, we consider the flooding schedule to simultaneously propagate messages on a general loopy graph, and view one round of message passing by eq \ref{eq:message1}-\ref{eq:message2} as one iteration in the recurrent computation~\cite{Zheng2015ConditionalNetworks}. We implement the iterated propagation in the message passing layer as unrolled propagations, which we illustrate in Fig~\ref{fig:message-passing-layer}.

The input to the message passing layer is potential functions given by the previous layers. The potential functions must produce positive values, and in this paper we suggest  a standard inner product with softplus activation function to model potentials~\cite{Tompson2014JointEstimation}. Given the previous layer output $\mathbf{z}$, we define the potential function by:
\begin{align}
  \phi_c(\mathbf{x}_c = X|\mathbf{z}) & \equiv \log \left(1 + \exp(\mathbf{w}_{X}^T\mathbf{z} + b_{X})\right). \label{eq:potential}
\end{align}

The first iteration of the message passing starts by setting $\mu^{(0)}_{\phi_c \rightarrow x}(x) = 1$, and the algorithm repeats eq \ref{eq:message1}-\ref{eq:message2} to propagate subsequent messages $ \mu^{(t)}_{x \rightarrow \phi_c}(x)$ and $\mu^{(t)}_{\phi_c \rightarrow x}(x)$ at step $t$. We apply unrolled iterations of message passing by the flooding schedule to the fixed number of iterations $T$. After iterated message propagations, we compute the approximate marginal probability (eq \ref{eq:marginal}). By unrolling iterations, we implement all the inference procedures in a feed-forward computation.

\subsection{Graph construction} \label{sec:graph-construction}
Although the sum-product algorithm can run inference on arbitrary graphical models, in practice it is computationally expensive to run an inference on dense factor graphs. The computational complexity is proportional to the factor degree and propagation steps, and also quadratic to the state space of a variable. Fortunately, in attribute recognition, the size of label space tends to be much smaller than pixel-based prediction, and sparse factor graphs work well in practice for attribute recognition.

In this paper, we suggest building a sparse graphical model for attributes based on variable coverage. We employ the following deterministic approach to build a factor graph: 1) compute correlation between attributes from training data; and 2) for each attribute, pick $k$ other attributes that has the highest absolute correlation, and form a potential over the pair. We empirically study the effect of different graph structures in Sec~\ref{sec:factor-graph-structure}.

\subsection{End-to-end learning} \label{sec:end-to-end}
The message passing layer in Sec \ref{sec:message-passing-layer} consists of sum, product, and normalization operations. We can run inference in a feed-forward manner in neural networks, and also apply the standard back-propagation for computing gradients. In this paper, we learn the attribute prediction model with the binary cross entropy loss:
\begin{align}
L &= \sum_{i} y_i \log p(x_i) + (1 - y_i) \log \left(1 - p(x_i) \right), \label{eq:loss}
\end{align}
where $y_i \in [0, 1]$ is the ground-truth label. We learn the model using SGD with momentum.

In practice, the gradient computation via back-propagation can be numerically unstable at normalization step in eq \ref{eq:message2} and \ref{eq:marginal}. A larger number of recurrent computation in the message passing layer results in vanishing gradients.
In Sec \ref{sec:experiments}, we only consider unary and pairwise potentials over binary variables, and apply two step propagation in the inference (repeating eq \ref{eq:message1}-\ref{eq:message2} two times).

\section{Experiments} \label{sec:experiments}

\subsection{Datasets} \label{sec:datasets}
In this paper, we use two datasets to evaluate the prediction performance: SUN Attribute dataset (SUN-Attr)~\cite{Patterson2014TheUnderstanding}, and CelebA dataset~\cite{liu2015faceattributes}. SUN-Attr contains 102 scene attributes in 14,340 images. Since the provided train-test partitions in SUN-Attr are not designed for joint training, we randomly split the images into 60\%/20\%/20\% train/val/test partitions for evaluation. CelebA contains 40 facial attributes in 202,599 images, and in this paper, we use the provided pre-cropped facial images with the original train/val/test splits for evaluation.

\subsection{Baselines} \label{sec:baselines}
We compare the following models in attribute prediction.

\noindent{\bf Sigmoid:} A standard binary classification model using a deep convolutional neural network that applies a linear function followed by a sigmoid function to produce the probability: $p(x_i=1) = \sigma(\mathbf{w}_i^T\mathbf{z}+b_i)$. We use ResNet-18 and ResNet-50 architecture~\cite{He2015} for feature extraction, and provide the average-pooled feature for the input $\mathbf{z}$ to the classifier. The sigmoid model is considered a baseline for independently predicting attributes~\cite{Zhang2014PANDA:Modeling}.\\
\noindent{\bf Const CRF:} Our structured prediction model that has softplus-linear unary potential (eq \ref{eq:potential}) and a softplus-constant pairwise potential:
\begin{align}
\phi_{i,j}(\mathbf{x}_{i,j} = X_{i,j}|\mathbf{z}) \equiv \log \left(1 + \exp(c_{X_{i,j}})\right).
\end{align}
The constant potential adds a static bias to the pairwise relationships as attempted in \cite{Chen2015LearningModels}. This model configuration has the effect of balancing prediction confidence between attributes by prior.\\
\noindent{\bf Linear CRF:} Our proposed model that uses the softplus-linear function (eq \ref{eq:potential}) for both unary and pairwise potentials. The model can be interpreted as a combination of first and second order classifiers.

For both Const and Linear CRFs, we define unary potentials for all variables, and pairwise potentials using the \texttt{min} approach described in Sec~\ref{sec:factor-graph-structure} that guarantees minimum number of pairwise factors per variable. In the baseline performance comparison, we evaluate the models with either at least 2 or 8 pairwise factors per variable in CRFs, and report the performance with better accuracy. Note that it is computationally prohibitive to define factors on all possible pairs of attributes. We fix the number of sum-product propagation to 2 in both CRF models.

\subsection{Training details} \label{sec:training-details}
We use the standard SGD with momentum to minimize the loss. We set the initial learning rate to {\tt 1e-1}, set the momentum to {\tt 1e-4}, and apply data augmentation by random cropping and horizontal flipping. To compare the CRF models with the baseline sigmoid, we first train the feature representation using sigmoid model from the ImageNet pre-trained CNN by the following learning strategy. We learn only the last classification layer for the first 20 epochs, apply SGD updates to all layers for the next 20 epochs, then drop the learning rate to {\tt 1e-2} for additional 20 epochs. After feature pre-training, we fine-tune all models including the sigmoid with the same learning strategy, and report the test performance of the best model in terms of validation accuracy. We also report the performance when only the last classification layer is trained with ImageNet pre-trained CNN without end-to-end learning. All experiments are implemented in PyTorch framework, and conducted on a computer with 4 GPUs.

\subsection{Quantitative evaluation} \label{sec:quantitative}
Table \ref{tab:imagenet-performance} summarizes the baseline configurations and the prediction performance in terms of accuracy, average precision, average recall, and average F1 score on all attributes for two datasets. We show the performance both when the classification layer (sigmoid or CRF) is trained with ImageNet pre-trained CNN and when the end-to-end learning strategy is employed. As a reference, we also show the performance of Torfason et al.~\cite{torfasonface} and Liu et al.~\cite{liu2015faceattributes} for CelebA dataset for state-of-the-art comparison.
\begin{table*}[t]
  \centering
  \caption{Model configurations and attribute prediction performance.}
  \label{tab:imagenet-performance}
  \begin{tabular}{|l|ll|cccc|cccc|}
    \hline
    & & & \multicolumn{4}{|c|}{ImageNet pre-trained} & \multicolumn{4}{|c|}{End-to-end trained} \\
    Dataset & CNN & Classifier & acc & avg pre & avg rec & avg F1 & acc & avg pre & avg rec & avg F1 \\
    \hline
    SUN-Attr & ResNet18 & Sigmoid    & 0.9512 & 0.6494 & 0.1961 & 0.2521 & 0.9547 & 0.6130 & 0.2797 & 0.3406 \\
             &          & Const CRF  & 0.9511 & {\bftab 0.6807} & 0.1794 & 0.2216 & 0.9547 & {\bftab 0.6427} & 0.2782 & 0.3343 \\
             &          & Linear CRF & {\bftab 0.9524} & 0.6312 & {\bftab 0.2472} & {\bftab 0.3042} & {\bftab 0.9554} & 0.6123 & {\bftab 0.3457} & {\bftab 0.4126} \\
             \cline{2-11}
             & ResNet50 & Sigmoid    & 0.9520 & 0.6803 & 0.2034 & 0.2638 & 0.9559 & 0.6531 & 0.3375 & 0.4105 \\
             &          & Const CRF  & 0.9524 & {\bftab 0.7190} & 0.1969 & 0.2461 & 0.9557 & {\bftab 0.6681} & 0.3113 & 0.3723 \\
             &          & Linear CRF & {\bftab 0.9538} & 0.6581 & {\bftab 0.2706} & {\bftab 0.3335} & {\bftab 0.9564} & 0.6280 & {\bftab 0.3735} & {\bftab 0.4418} \\
    \hline
    CelebA   & ResNet18 & Sigmoid    & 0.8667 & 0.6712 & 0.4333 & 0.4927 & 0.9152 & {\bftab 0.7843} & 0.6846 & 0.7200 \\
             &          & Const CRF  & 0.8771 & 0.6926 & 0.4415 & 0.5028 & {\bftab 0.9192} & 0.7734 & 0.7009 & 0.7280 \\
             &          & Linear CRF & {\bftab 0.8799} & {\bftab 0.6937} & {\bftab 0.4642} & {\bftab 0.5253} & 0.9187 & 0.7691 & {\bftab 0.7065} & {\bftab 0.7294} \\
             \cline{2-11}
             & ResNet50 & Sigmoid    & 0.8733 & 0.6910 & 0.4458 & 0.5031 & 0.9142 & 0.7779 & 0.6885 & 0.7172 \\
             &          & Const CRF  & 0.8820 & {\bftab 0.6967} & 0.4699 & 0.5229 & {\bftab 0.9191} & 0.7767 & {\bftab 0.6994} & {\bftab 0.7256} \\
             &          & Linear CRF & {\bftab 0.8838} & 0.6965 & {\bftab 0.4947} & {\bftab 0.5447} & 0.9190 & {\bftab 0.7801} & 0.6888 & 0.7219 \\
             \cline{2-11}
             & \multicolumn{2}{|l|}{ANet+LNet~\cite{liu2015faceattributes}}       &        &        &        &        & 0.8700 &        &        &  \\
             & \multicolumn{2}{|l|}{Torfason et al.~\cite{torfasonface}}       &        &        &        &        & 0.9022 &        &        &  \\
    \hline
  \end{tabular}
\end{table*}

In overall, introducing structured inference improves the prediction performance over the baseline sigmoid classifiers in almost all cases. The Linear CRF consistently improves accuracy, average recall, and average F1 in SUN-Attr, with a solid performance advantage on the average F1 measure. Const CRF models also show advantage in average precision in SUN-Attr. Linear CRFs outperform Const CRFs in accuracy in most cases, except for the on-par performance observed in the end-to-end trained cases in CelebA.

\paragraph*{End-to-end learning}
In CelebA dataset, CRF models outperform the baseline sigmoid classifier, but the significant improvement is in between the ImageNet pre-trained models and the end-to-end learned models. The result suggests that in CelebA the internal representation is significantly more important than learning the inter-attribute relationship due to that ImageNet has more similar image statistics to SUN dataset (scene images) than CelebA (facial images). As a result, in SUN-Attr, the effect of inter-dependent prediction becomes more apparent.

The difference between ResNet18 and ResNet50 shows the effect of having a richer feature representation in CNN. The effect seems clear in CelebA pre-trained cases, as the performance difference diminishes after end-to-end training. The result confirms that ImageNet pre-trained CNN is not optimized for CelebA dataset.

\begin{figure*}[t]
  \includegraphics[width=\textwidth]{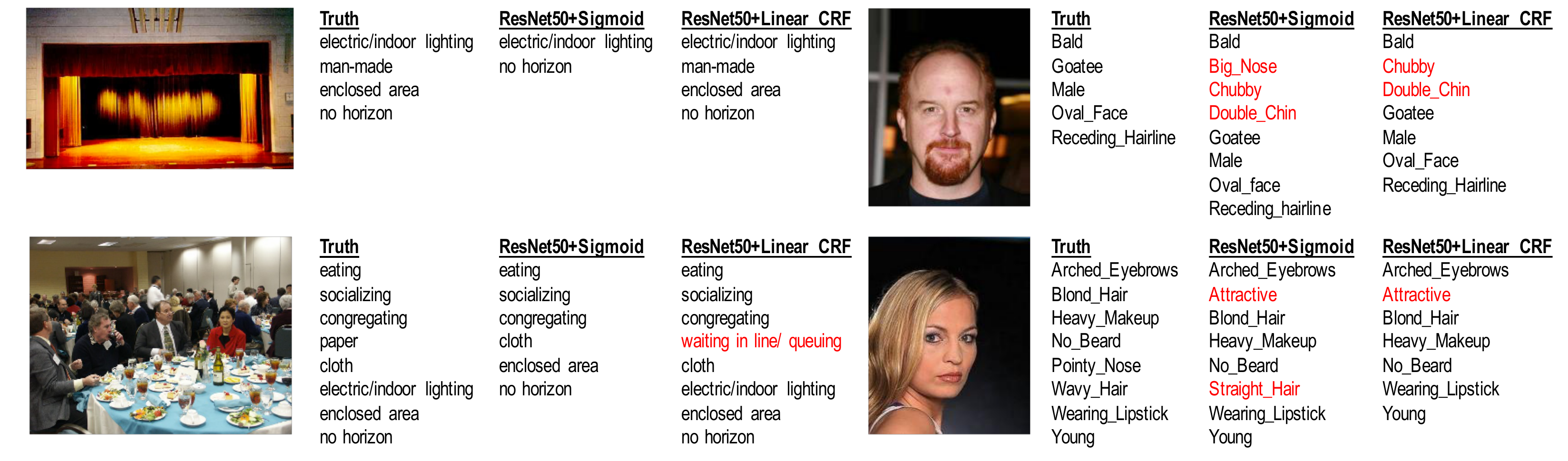}
  \caption{Attribute prediction examples (Left col.: SUN-Attr dataset, right col.: CelebA dataset).  False positives are marked red.}
  \label{fig:predictions}
\end{figure*}

\paragraph*{Choice of potential functions}
The Const CRF results in worse average F1 to the sigmoid baseline in SUN-Attr dataset. We suspect that the learned prior in potentials somewhat makes more conservative prediction (high precision) under limited training data.

The improvement by the Linear CRF model over the sigmoid baseline looks interesting, because we are predicting second-order relationship in addition to each attribute in the deep neural networks; \emph{i.e.}, predicting a pair $(x_i, x_j) \in \{(0, 0), (0, 1), (1, 0), (1, 1)\}$. The performance improvement by the Linear CRF clearly indicates that the second-order prediction learns the structure information in the inter-attribute relationship that an independent baseline (sigmoid) cannot capture in the inference.

\paragraph*{Structured inference or internal representation?}
One interesting observation is that when we apply end-to-end learning, the difference between the baseline sigmoid and the CRF models becomes smaller (e.g., SUN-Attr accuracy). The result suggests a deep neural network can learn a good internal representation $\bf z$ that also encodes inter-attribute relationships in the representation itself. We can observe the significance of the structured inference by comparing the improvement by CRF models over the sigmoid models and the improvement by ResNet50 over ResNet18. In CelebA, the difference between ResNet18 and ResNet50 diminishes after end-to-end training, but we still observe some advantage by the structured inference in both CRF models. We believe the structured inference is capturing inter-attribute relationships that a feature representation is unable to model even after end-to-end training.


\subsection{Qualitative analysis} \label{sec:qualitative}
Fig.~\ref{fig:predictions} shows a few prediction examples for the two datasets using ResNet50 + Sigmoid or Linear CRF models after end-to-end learning. Predictions are mostly similar, but Linear CRF tends to give more coverage, as indicated by higher recall in Table \ref{tab:imagenet-performance}.

When the inter-attribute correlation matters, our CRF models tend to complement the missing attributes in the baseline prediction. For example, in the upper-left of Fig~\ref{fig:predictions}, the Linear CRF prediction correctly infer \emph{man-made} and \emph{enclosed-area} attribute knowing that the scene is indoor without a horizon.

A mistake happens typically when the attributes are not obvious from the image. For example, whether the face is \emph{attractive} involves highly subjective judgment in CelebA, and the annotation might not be absolutely correct. Both our models and the baseline sigmoid tend to make a mistake in such attributes.


\section{Ablation studies} \label{sec:ablation-studies}
In this section, we empirically study the effect of various architectural configurations using ResNet18 CNN and SUN-Attr dataset. 

\begin{figure*}[t]
  \centering
  \includegraphics[width=\textwidth]{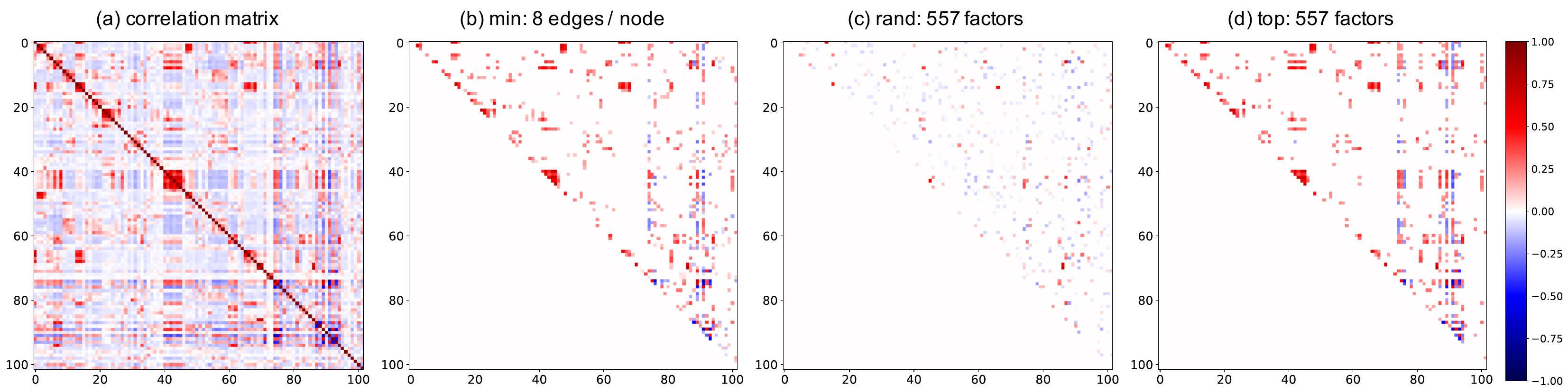}
  \caption{Correlation matrix in SUN-Attr dataset (a) and adjacency matrices for a factor graph corresponding to \texttt{min}, \texttt{rand}, and \texttt{top} policies (b-d). Color indicates correlation coefficients. The \texttt{min} and \texttt{top} policy looks similar, but guaranteeing edges in every node (b) or scattering edges (c) improves the prediction performance than naively taking highly correlated pairs (d).}
  \label{fig:correlation-factors}
\end{figure*}

\subsection{Factor graph structure} \label{sec:factor-graph-structure}
What is the effective sparse graph structure for attributes in the given dataset? To answer this question, we compare the performance of our model under three graph construction policies.\\
\noindent\textbf{min}: Minimum $K$ edge-per-node policy as in Sec~\ref{sec:graph-construction}. For each attribute, we pick $K$ highest-correlated other attributes in the training data, and form a pair. This policy guarantees at least $K$ pairwise factors exist for each attribute.\\
\noindent\textbf{rand}: Random factor policy, where we pick random $N$ pairs of attributes and form a pairwise factor.\\
\noindent\textbf{top}: Top $N$ correlation policy. We compute the correlation matrix from the training data, and pick $N$ pairs of attributes according to the order of highest absolute correlation. This policy is similar to the positive and negative correlation approach employed in Hu et al~\cite{Hu2016LearningRelations}.

\begin{figure}[t]
  \centering
  \includegraphics[width=\columnwidth]{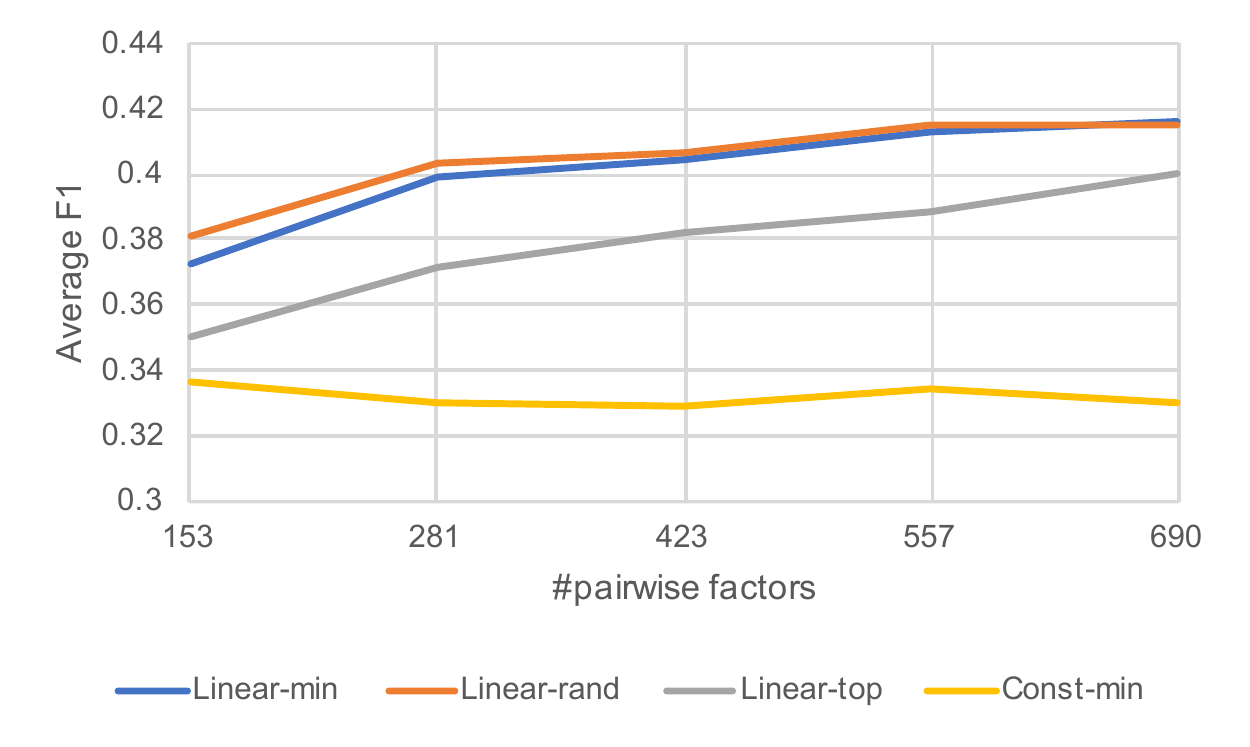}
  \caption{Average F1 score vs. the size of pairwise factors using ResNet18 + CRF models in SUN-Attr dataset. We compare \texttt{min}, \texttt{rand}, and \texttt{top} policies to construct a factor graph.}
  \label{fig:f1-factor}
\end{figure}
We evaluate the average F1 score for different number of pairwise factors in the graph. For \texttt{min} policy, we evaluate the performance at $K=2,4,6,8,10$, which result in the pairwise factor size $N=153,281,423,557,690$ respectively. For \texttt{rand} and \texttt{top} policy, we choose the same number of factors in the graph. This condition measures how a different factor graph of the equivalent size affects the prediction performance. Note that it is computationally infeasible to learn a fully-connected factor graph for SUN-Attr dataset. 
We pick Linear CRF for evaluating the above three policies as well as Const CRF with \texttt{min} policy for reference. Fig \ref{fig:correlation-factors} visualizes the correlation matrix of SUN-Attr dataset and the adjacency matrix of the factor graphs for \texttt{min} with $K=8$ and \texttt{rand}, \texttt{top} with $N=557$ cases. The \texttt{min} policy looks similar to the graph structure from the \texttt{top} policy, but the \texttt{top} policy tends to assign larger number of factors to smaller number of variables.

Fig \ref{fig:f1-factor} plots the average F1 score over different number of pairwise factors in the graph. We fix the number of iterations in message passing to 2 in this experiment. We observe that F1 score in general increases as the number of pairwise factors increases in Linear CRF, but there is no benefit in Const CRF. The inferior Const CRF performance suggests that static bias on pairs have minor effect in prediction, even if we consider larger number of pairs.

The more interesting observation is that the \texttt{top} policy is consistently worse than the other two policies. The difference between \texttt{min} and \texttt{rand} is small, though \texttt{rand} introduces randomness to the model and is less interpretable. The result implies that it is better to introduce higher-order prediction to as many variables as possible than choosing structure solely based on correlation.
We suspect two reasons for this observation. The first is that highly correlated attributes are rather easy to predict than not-so-obvious pairs. For example, \emph{open area} and \emph{natural light} relationship is obvious and a CNN can already learn a good internal representation to predict both through end-to-end learning. However, when we have an explicit potential function on a not obvious pair such as \emph{foliage} and \emph{vacationing}, we might have more chance to uncover a hidden relationship between them from the dataset. The second reason is numerical stability in message passing. In eq~\ref{eq:message2}, we need to take the product of many messages, and possibly having many factors concentrated on one variable leads to unstable inference and back propagation in sum-product algorithm.

Although the larger number of pairwise factors help improve the prediction, there is a drawback in computational complexity. Sec~\ref{sec:computational-complexity} empirically analyzes runtime. The choice of the higher-order factor size should be subject to the application requirement for speed and quality.


\subsection{Message propagation depth} \label{sec:propagation-depth}

\begin{figure}[t]
  \centering
  \includegraphics[width=\columnwidth]{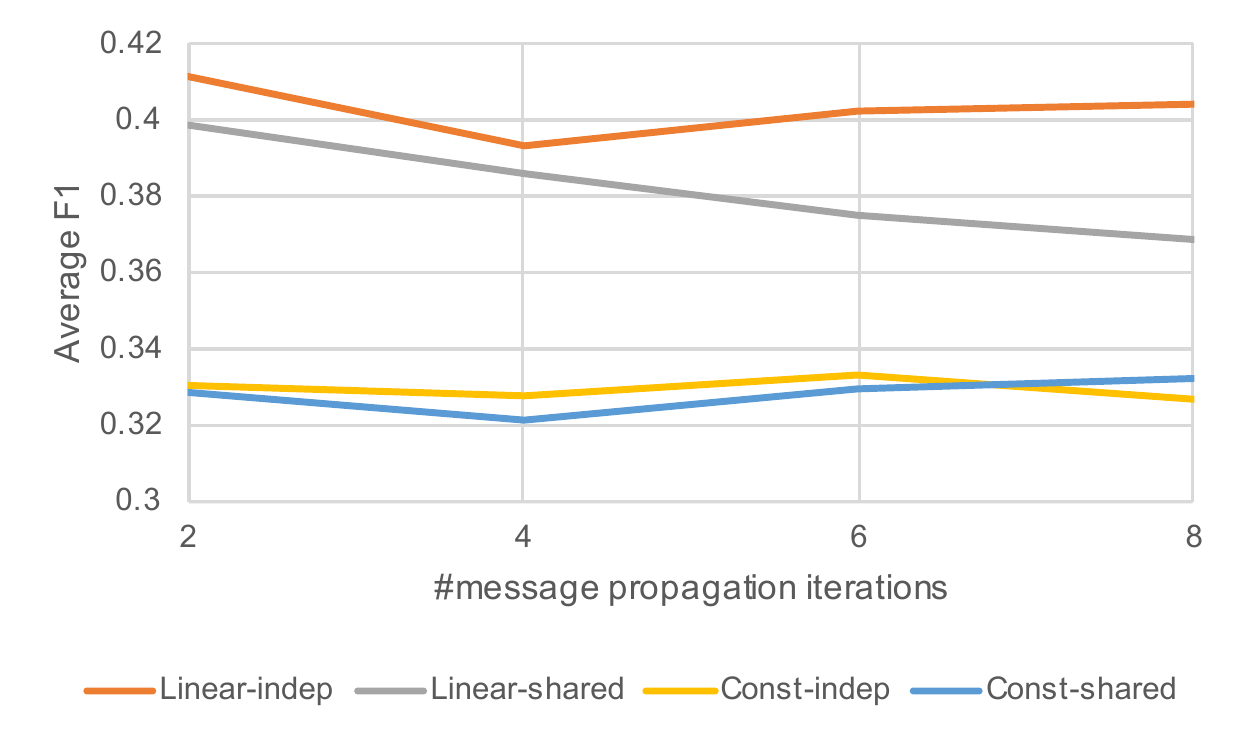}
  \caption{Average F1 score vs. the number of message iterations using ResNet18 + CRF models in SUN-Attr dataset. We show independent potentials and shared potentials for both Const and Linear CRF models.}
  \label{fig:f1-iterations}
\end{figure}

Our message passing layer requires repeated message propagations by flooding schedule (eq \ref{eq:message1} and \ref{eq:message2}). How many times do we need to iterate message propagation? We report the prediction performance with respect to the number of iterated propagations. Fig~\ref{fig:f1-iterations} plots average F1 score over the number of message propagations for 4 different models, all with \texttt{min-4} policy factor configurations and independently learned. The \textbf{Linear-shared} and \textbf{Const-shared} plots indicate the performance of Linear CRF and Const CRF respectively. As the plot suggests, the Linear CRF degrades the average F1 with increased number of iterations, while Const CRF keeps similar score regardless of the iterations. We suspect that as we increase the number of propagations in inference, parameter learning via back-propagation suffers more from the vanishing gradient problem and results in inferior performance. Increased iterations also indicate increased computational overhead. Although we might be able to circumvent the vanishing gradient problem (e.g., by \cite{Ioffe2015BatchShift}), there seems no practical benefit in increasing the depth of the message propagation layer in our experimental setup.

\subsection{Independent factors in propagation} \label{sec:independent-factors}
Our model implements message passing procedure by unrolling the iterated message passing (eq \ref{eq:message1} and \ref{eq:message2}). It is possible to make the potential function independent in each message passing iteration, as attempted in
the previous work on inference machine~\cite{Ross2011LearningPrediction,Ramakrishna2014PoseMachines}. Here, we examine the effect of independently modeling potential functions at each step in the flooding sum-product propagation.

The \textbf{Linear-indep} and \textbf{Const-indep} plots in Fig~\ref{fig:f1-iterations} show the performance of Linear CRF and Const CRF with \texttt{min-4} policy graph and independent potentials per iteration. We observe that the independent potentials slightly outperform the shared potentials in Linear CRF (\textbf{Linear-indep} vs \textbf{Linear-shared}), but the performance is almost the same in Const CRF. The number of iterations no longer degrades the average F1 score in Linear CRF. We suspect that with independent potential functions, potential functions in the later propagation step avoid vanishing gradients that happen to potentials in earlier steps and are able to learn parameters. When parameters are shared across iterations, vanishing gradients to earlier steps prevents proper end-to-end learning of potentials. In any case, as in the previous section, there is a computational overhead in increased message propagation with no performance benefit. We conjecture that the \emph{width over depth} strategy is the best to employ with sum-product message passing in the deep networks.

\subsection{Model size and runtime} \label{sec:computational-complexity}
\begin{figure}
  \centering
  \includegraphics[width=\columnwidth]{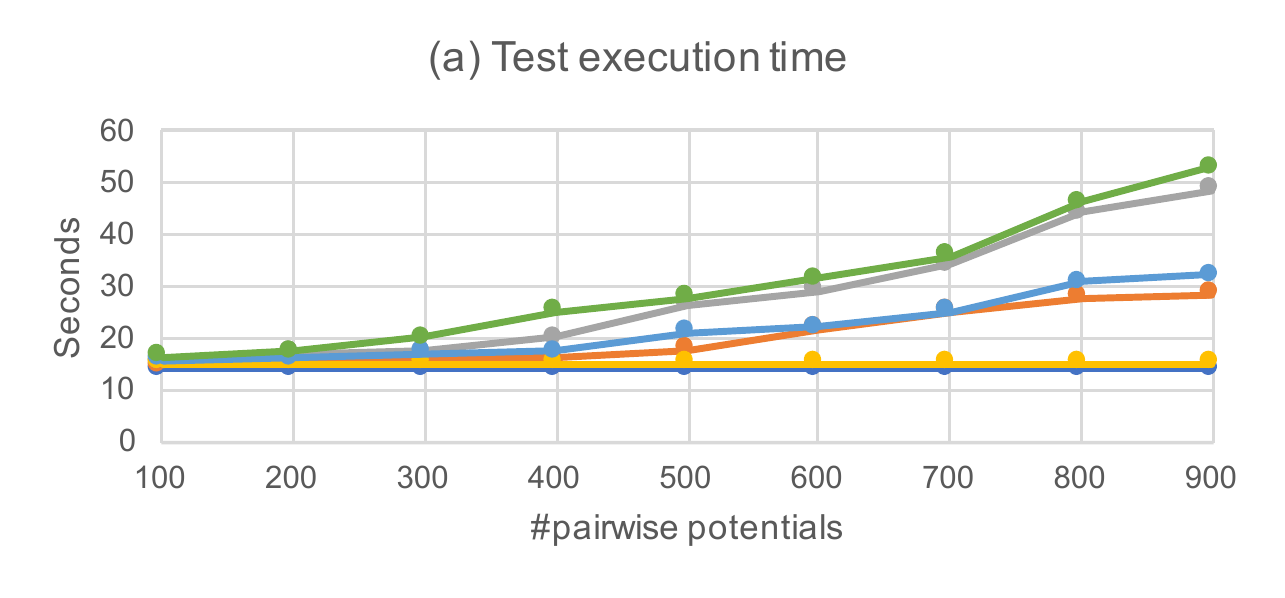}
  \includegraphics[width=\columnwidth]{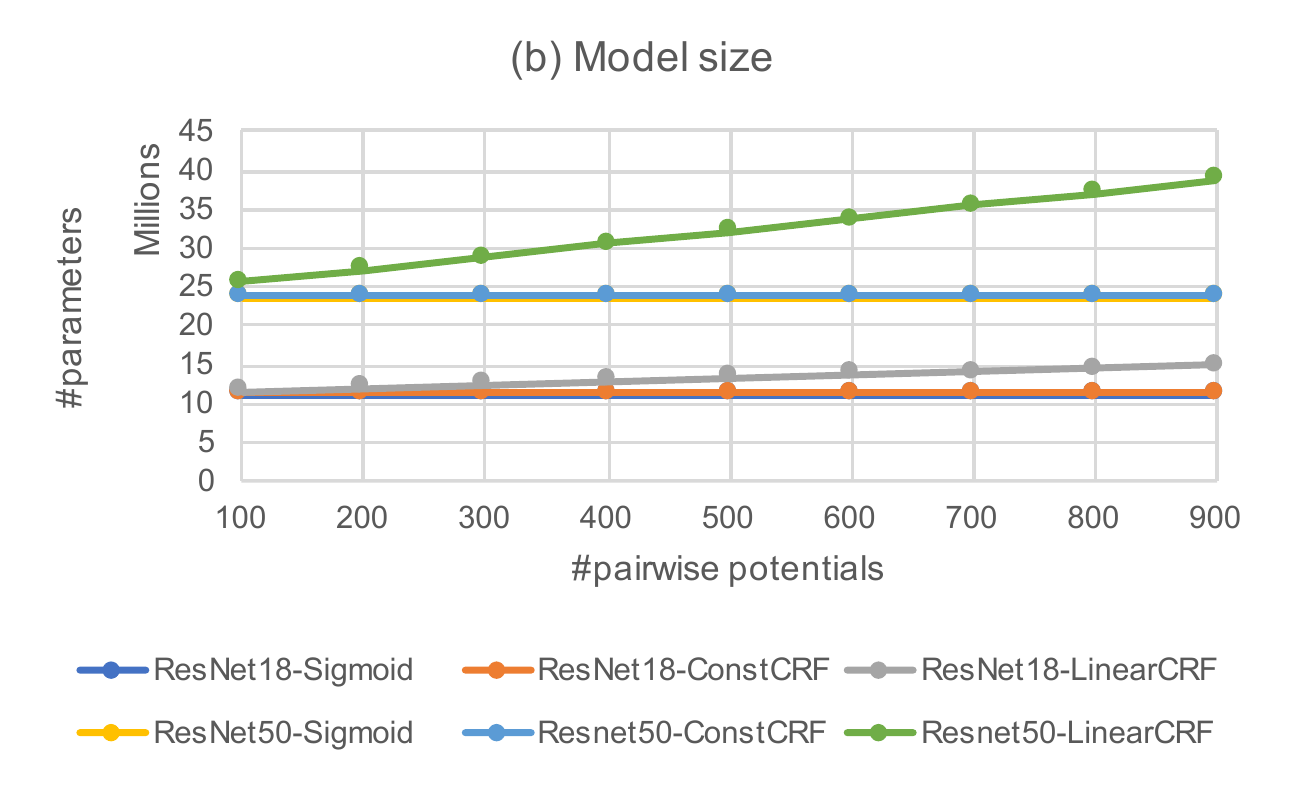}
  \caption{(a) Test execution time in seconds and (b) model parameter size with respect to the size of pairwise potential functions in the model.}
  \label{fig:runtime-modelsize}
\end{figure}
What is the computational cost to introduce sum-product inference? We evaluate the time to run inference on 2,868 test images in SUN-Attr dataset with respect to the number of pairwise potentials in the graphical model. Fig~\ref{fig:runtime-modelsize}a plots the runtime of sigmoid baseline, Const CRF, and Linear CRF using \texttt{rand} policy to construct $N=100, 200, \cdots, 900$ pairwise potentials. Fig~\ref{fig:runtime-modelsize}b plots the parameter size of each model. We compare both ResNet18 and ResNet50 architecture. The runtime is evaluated on a workstation with 2 Intel Xeon 1.90GHz CPUs and 1 NVIDIA Titan-X (Pascal) GPU. Runtime includes disk IOs where we use batch size equal to 256.

The plot indicates that as the number of pairwise potential increases, sum-product inference dominates the computation time. Computational cost of Linear CRF roughly doubles to the baseline sigmoid when 600 pairwise factors are introduced, which is equivalent to around 8 edges per variable under \texttt{min} policy graph construction in Sec~\ref{sec:factor-graph-structure}. We find that ResNet18 + Linear CRF (13.7M parameters) performs better than or on par to ResNet50 + sigmoid (23.7M parameters) that has a larger number of parameters.

Although Linear CRFs give performance benefit, there is certainly a drawback in computational cost. Linear CRFs quickly increase the parameter size as we introduce more potential functions, since the size of softplus-linear potential function is proportional to the size of the feature input. The execution time is not proportional to the parameter size in this paper, because our implementation is not yet highly optimized for parallelism. The bottleneck is in iterated message propagation where our implementation computes messages in a sequential loop. Parallelizing message computation could potentially significantly boost the execution. It is our future work to further optimize the message passing layer. In any case, per-image processing time is in the order of 0.01 seconds in all models and practical for various applications.

\section{Conclusion and future work} \label{sec:conclusion}

We have presented the structured inference approach for multiple attribute prediction. Our framework is based on the marginal inference in the conditional random field, and we propose to model the potential functions by deep neural networks. The empirical study using SUN-Attr and CelebA datasets suggest that the structured inference improves the attribute prediction performance. Detailed analysis on factor size, message depth, and runtime reveals the characteristics of our message passing layer and also suggests the practical design principle for factor graph construction.

In this paper, we have not evaluated higher-order potential functions with more than 2 variables. It would be interesting to see learning higher-order potential by back-propagation~\cite{Arnab2016HigherNetworks}, while mitigating the computational complexity of the sum-product algorithm.
Our approach of using sparse graphical models with sum-product inference as a drop-in replacement for classifiers works well for attribute recognition. We wish to investigate if a similar approach works for other inference problems. Also scaling up the model to larger attributes is another future work~\cite{Deng2014Large-scaleGraphs}.

\subsection*{Acknowledgements} \label{sec:acknowledgements}
This work was supported by JSPS KAKENHI JP16H05863 and by CREST, JST JPMJCR14D1.

\newpage
{\small
\bibliographystyle{ieee}
\bibliography{references}
}

\end{document}